\documentclass[10pt,twocolumn,twoside]{IEEEtran}           % TWOCOLS

\usepackage[usenames]{color}
\usepackage[ansinew]{inputenc}
\usepackage{inputenc}
\usepackage{graphicx}
\usepackage{amsfonts}
\usepackage{amssymb}
\usepackage{amsbsy}
\usepackage{amsmath}
\usepackage{multirow}
\usepackage{multicol}
\usepackage{array}
\usepackage{color}
\usepackage{soul}
\usepackage{cite}
\usepackage[normalem]{ulem} % either use this (simple) or
\usepackage{times}
\usepackage[all]{xy}
\usepackage{lscape}
\usepackage{flushend}
\usepackage{url}
\usepackage{float}
\usepackage{rotating}
\usepackage{adjustbox}
\usepackage{enumerate}
\usepackage{verbatim}

\usepackage[position=top]{subfig}

\newcommand{\Real}{\mathbb R}

\def\x{{\mathbf x}}

\newcommand{\vect}[1]{{\boldsymbol{\mathbf{#1}}}} % vector
\newcommand{\mat}[1]{{\boldsymbol{\mathbf{#1}}}} % matrix
\newcommand{\y}{{\vect y}}

\newcommand{\Kf}{{\mat K}_\vect{ff}}

\newcommand{\dataset}{{\cal D}}

\newcommand{\GP}[0]{\mathcal{GP}} 

\newcommand{\Normal}{\mathcal{N}} 
 
\newcommand{\prob}{p}
 
\newcommand{\cut}[1]{} % cut out a part of the text

\usepackage{hyperref}
\hypersetup{
    unicode=false,          % non-Latin characters in Acrobat’s bookmarks
    pdftoolbar=true,        % show Acrobat’s toolbar?
    pdfmenubar=true,        % show Acrobat’s menu?
    pdffitwindow=false,     % window fit to page when opened
    pdfstartview={FitH},    % fits the width of the page to the window
    pdftitle={WGP},         % title
    pdfauthor={Anna Mateo}, % author
    pdfsubject={Warped GPs} % subject of the document
    pdfcreator={}           % creator of the document
    pdfproducer={},         % producer of the document
    pdfkeywords={},         % list of keywords
    pdfnewwindow=true,      % links in new window
    colorlinks=true,        % false: boxed links; true: colored links
    linkcolor={blue},       % color of internal links (change box color with linkbordercolor)
    citecolor=blue,         % color of links to bibliography
    filecolor=blue,         % color of file links
    urlcolor=blue           % color of external links
}

\newcommand{\blue}[1]{\textcolor[rgb]{0,0,0.0}{#1}}         % Color black!

\begin{document}

%\title{\blue{Biophysical Parameter Retrieval with Warped Gaussian Processes}}  % IGARSS 
\title{Warped Gaussian Processes in Remote Sensing Parameter Estimation and Causal Inference}

\author{
Anna Mateo-Sanchis,
Jordi Mu\~noz-Mar\'i,
Adri\'an P\'erez-Suay,
Gustau Camps-Valls
\thanks{This work was funded by the European Research Council (ERC) under the ERC-CoG-2014 SEDAL project (grant agreement 647423), and the Spanish Ministry of Economy, Industry and Competitiveness (MINECO) under the `Network of Excellence' program (grant code TEC2016-81900-REDT), and MINECO/ERDF under CICYT projects TIN2015-64210-R and TEC2016-77741-R.}
\thanks{Authors are with the Image Processing Lab (IPL), Universitat de Val\`encia, Spain. E-mail: [anna.mateo,jordi.munoz,adrian.perez,gustau.camps]@uv.es, \url{http://isp.uv.es}. } %Vicarious.com, San Francisco, USA}
}

\maketitle

\begin{abstract}
This paper introduces warped Gaussian processes (WGP) regression in remote sensing applications. WGP models output observations as a parametric nonlinear transformation of a GP. The parameters of such prior model are then learned via standard maximum likelihood. 
We show the good performance of the proposed model for the estimation of oceanic chlorophyll content from multispectral data, vegetation parameters (chlorophyll, leaf area index, and fractional vegetation cover) from hyperspectral data, and in the detection of the causal direction in a collection of 28 bivariate geoscience and remote sensing causal problems. The model consistently performs better than the standard GP and the more advanced heteroscedastic GP model, both in terms of accuracy and more sensible confidence intervals.
%focuses on biophysical parameter retrieval based on Gaussian Processes (GPs). Very often an arbitrary transformation is applied to the observed variable (e.g. chlorophyll content) to better pose the problem. This standard practice essentially tries to linearize/uniformize the distribution by applying non-linear link functions like the logarithmic, the exponential or the logistic functions. In this paper, we propose to use a GP model that automatically learns the optimal transformation directly from the data. 
%The so-called warped GP regression (WGP) presented in~\cite{Snelson03} models output observations as a parametric nonlinear transformation of a GP. The parameters of such prior model are then learned via standard maximum likelihood. We show the good performance of the proposed model for the estimation of oceanic chlorophyll content, which outperforms the regular GP and a more advanced heteroscedastic GP model.
\end{abstract}


\section{Introduction} \label{sec:intro}

\IEEEPARstart{W}{ith} the forthcoming superspectral satellite missions dedicated to land, vegetation and ocean monitoring, an unprecedented data stream is now available and will increase in following years. Automatic efficient techniques for spatial and temporal explicit quantification of Earth properties are an urgent need. When it comes to the implementation of candidate retrieval methods into operational data processing chains, like e.g. the Copernicus' Sentinels, it is mandatory to invest in models that are both accurate and robust, but also requiring minimum user intervention for fitting parameters, and that provide sensible confidence intervals for the predictions. This is the scenario where this paper is placed. 

In this paper, we focus on the Bayesian non-parametric framework in general, and in Gaussian processes (GPs)~\cite{Rasmussen2006}, which have yielded very good performance in the last years in many geoscience and remote sensing problems, such as biophysical parameter estimation, radiative transfer model emulation and causal inference from empirical data~\cite{Verrelst12,CampsValls16grsm}. %Verrelst12jstars,
Despite the very good performance of standard GPs, inclusion of prior knowledge respecting signal characteristics is mandatory to achieve state-of-the-art results, and in turn make parameter tuning simpler and less sensitive to initializations. This can be achieved by designing kernel functions that respect signal smoothness in space or time~\cite{DuvLloGroetal13,Sancho14}, combining multi-scale and multi-resolution of the features~\cite{Dunson12,Sancho14}, %by encoding invariances to distortions of the input space~\cite{izquierdo13}, 
or by designing regularizers to encompass signal and noise relations~\cite{CampsVallsGRSL2013}. 

Little attention has been paid, however, to the statistical characteristics of the output, observed variable. 
Actually, we should note an important observation: GP modeling assumes that the target data is distributed as a multivariate Gaussian and that observations are buried in \blue{constant power} (homoscedastic) Gaussian noise as well. These assumptions allow model tractability, yet sometimes they are unrealistic in practice.
Commonly one performs {\em ad hoc} pre-processing of the data to achieve transformations of the observed variable to make it look as a Gaussian. 
It is customary to apply logarithmic, exponential, power or logistic transformations to spectral ratios. This is not only arbitrary, but somehow contradictory since first data is transformed by a parametric function, and then, a non-parametric GP model is fitted.
In this paper, we propose a GP that automatically learns the optimal transformation to be applied.
The method is called warped GP (WGP)~\cite{Snelson03} and commonly leads to more accurate results over standard and more advanced GPs. It also leads to more sensible confidence intervals and gives important insights on the non-linearity of the problem. The WGP model actually generalizes standard GPs, and allows to deal with non-Gaussian processes and non-Gaussian noise.

The remainder of the paper is organized as follows. Section 2 reviews the warped GP regression method. Section 3 details the data characteristics used in this paper. Section 4 gives the experimental results for several problems: (i) the estimation of chlorophyll-a concentration from remote sensing upwelling radiance just above the ocean surface, estimation of biophysical parameters (chlorophyll-a, leaf area index, and fractional vegetation cover) from hyperspectral data, and the problem of cause-effect discovery from pairs of observational data. Finally, Section 5 concludes the paper and outlines the further work. 

%%%%%%%%%%%%%%%%%%%%%%%%%%%%%%%%%%%%
%%%%%%%%%%%%%%%%%%%%%%%%%%%%%%%%%%%%
%%%%%%%%%%%%%%%%%%%%%%%%%%%%%%%%%%%%
%%%%%%%%%%%%%%%%%%%%%%%%%%%%%%%%%%%%
\section{Warped Gaussian Processes}\label{sec:wgp}

This section reviews the theory of GP regression, introduces the warped GP model, and describes the parameterization of both the covariance and the warping functions.

%\subsection{Regression and function approximation}
%Estimation, regression and function approximation are old, largely studied problems in statistics and machine learning. The problem boils down to optimize a loss (cost, energy) function over a class of functions. A large class of regression problems in particular are defined as the joint minimization of a loss function accounting for errors of the function $f\in {\mathcal H}$ to be learned, and a regularization term, $\Omega\left(\|f\|^2_{{\mathcal H}}\right)$, that controls its capacity (excess of flexibility). The problem can be approached within a Bayesian nonparametric framework, and several algorithms are available, such as the relevance vector machine (RVM)~\cite{Tipping2001a,campsvalls06rvm} or Gaussian Process (GP)~\cite{Rasmussen2006,Verrelst12}, in which we will focus here.

%%%%%%%%%%%%%%%%%%%%%%%%%%%%%%%%%%%%
\subsection{Gaussian Process Regression (GP)}

Standard regression approximates observations (often referred to as \emph{outputs}) $\{y_i\}_{i=1}^{N}$ as the sum of some unknown latent function $f(\x)$ of the inputs $\{\x_i \in\Real^D\}_{i=1}^{N}$ plus \emph{constant power} Gaussian noise, that is:
\begin{equation}
    y_i = f(\x_i) + \varepsilon_i,~~~~\varepsilon_i \sim\Normal(0,\sigma_n^2). 
\end{equation}
Instead of proposing a parametric form for $f(\x)$ and learning its \blue{parameters} in order to fit observed data well, GP regression proceeds in a Bayesian, non-parametric way. A zero mean\footnote{It is customary to subtract the sample mean to data, and then to assume a zero mean model.} 
GP prior is placed on the latent function $f(\vect{x})$ and a Gaussian prior is used for each latent noise term $\varepsilon_i$, 
$f(\vect{x})\;\sim\;\GP(\vect{0}, k_\vect{\theta}(\vect{x},\vect{x}'))$, 
where $k_\vect{\theta}(\vect{x},\vect{x}')$ is a covariance function parameterized by a vector $\vect{\theta}$, and $\sigma_n^2$ is a parameter accounting for the noise power.
Essentially, a Gaussian process is a stochastic process whose marginals are distributed as a multivariate Gaussian. In particular, given the priors $\GP$, samples drawn from $f(\x)$ at the set of locations $\{\x_i\}_{i=1}^N$ follow a joint multivariate Gaussian with zero mean and covariance matrix $\mat{K_\vect{ff}}$ with  $[\mat{K_\vect{ff}}]_{ij} = k_\vect{\theta}(\vect{x}_i,\vect{x}_j)$.

If we consider a test location $\x_*$ with corresponding output $y_*$, the $\GP$ defines a joint %the following  joint 
prior distribution between the observations $\y \equiv \{y_i\}_{i=1}^N$ and $y_*$.
\iffalse
\begin{equation*}
  \left[\!\! 
    \begin{array}{c}
      \vect{y} \\
      y_*
    \end{array}
    \!\!\right]
  \;\sim\;
  \Normal\left( \vect{0},\;\left[\!\!
      \begin{array}{cc}
        \mat{K}_{\vect{ff}}+\sigma_n^2\mat{I} & \vect{k}_{\vect{f}*}\\
        \vect{k}_{\vect{f}*}^\top & k_{**}+\sigma_n^2\\
      \end{array}
      \!\!\right]\right).
      \label{eq:jointprior}
\end{equation*}
\fi
Collecting available data in $\dataset\equiv\{(\vect{x}_i,y_i)|i=1,\ldots
N\}$, it is possible to analytically compute the posterior distribution over the unknown output $y_*$: 
%
%\begin{subequations}
%\label{eq:preddist} 
\begin{equation}
\begin{split}
    %\label{eq:preddista}  
 \prob(y_*|\vect{x}_*,\dataset)&=\Normal(y_*|\mu_{\text{GP}*},\sigma_{\text{GP}*}^2)\\
%\label{eq:preddistb}  
\mu_{\text{GP}*} &= \vect{k}_{\vect{f}*}^\top (\mat{K}_{\vect{ff}}+\sigma_n^2\mat{I})^{-1}\vect{
y} = \vect{k}_{\vect{f}*}^\top\boldsymbol{\alpha} \\
%\label{eq:preddistc}  
\sigma_{\text{GP}*}^2 &= \sigma_n^2+k_{**}-
     \vect{k}_{\vect{f}*}^\top (\mat{K}_{\vect{ff}}+\sigma_n^2\mat{I})^{-1}\vect{k}_{\vect
{f}*}.
\end{split}
\end{equation}
%\end{subequations}
%which is computable in $\bigO(n^3)$ time (this cost arises from the inversion of the $n\times n$ matrix $\mat{K}_{\vect{ff}}+\sigma^2\mat{I}$), see \cite{Rasmussen2006}, Ch.~8.
The corresponding \blue{parameters} $\{\vect{\theta},\sigma_n\}$ are typically selected by Type-II Maximum Likelihood, using the marginal likelihood (also called evidence) of the observations, which is also analytic (explicitly conditioning on $\vect{\theta}$ and $\sigma_n$):
\begin{equation}
\log p(\vect{y}|\vect{\theta},\sigma_n) =  \log \Normal(\y|\vect{0}, \Kf + \sigma_n^2\mat{I}).
\label{eq:logevidence}
\end{equation}
%When the derivatives of \eqref{eq:logevidence} are also analytic, which is often the case, conjugated gradient ascend is typically used for optimization.

%%%%%%%%%%%%%%%%%%%%%%%%%%%%%%%%%%%%
\subsection{Warped Gaussian Process Regression (GP)}

In real applications, the distribution of the observations is very often not Gaussian, due to the sampling strategies followed during the protocols in data collection or because of the natural variability of the problem. Very often, in practice, one remedies that by transforming the observed variable to make it look like a Gaussian. Actually, it is a standard practice to apply logarithmic or exponential functions to this end.

In this paper, we use a GP model that automatically learns the optimal transformation by warping the predictions of a standard GP model. The method is called {\em warped} GP~\cite{Snelson03}, and essentially warps observations ${\bf y}$ through a nonlinear parametric function $g$ to a latent space:
$$z_i= g(y_i,\psi) = g(f({\bf x}_i) + \varepsilon_i),$$
where $g$ is a function with scalar inputs parameterized by $\vect{\psi}$. \blue{Note that the new error term cannot be readily accessed or quantified, since {\em g} is a nonparametric function and does not factorize.}  
The function $g$ must be {\em monotonic}, otherwise the probability measure will not be conserved in the transformation, and the distribution over the targets might not be valid~\cite{Snelson03}. It can be shown that replacing $y_i$ by $z_i$ into the standard GP model leads to an extended problem that can be solved by taking derivatives of the negative log likelihood function in~\eqref{eq:logevidence}, but now with respect to both $\vect{\theta}$ and $\vect{\psi}$ \blue{parameter} vectors. %The problem can be actually solved by using a conjugate gradient method to compute maximum likelihood parameter values.

%%%%%%%%%%%%%%%%%%%%%%%%%%%%%%%%%%%%
\subsection{Model parametrization}

Owing to the probabilistic treatment, all GP variants yield a full posterior predictive distribution over $y_*$, it is possible to obtain not only mean predictions for test data, but also its uncertainty. %The predictive mean is given by $\mu_{\text{GP}*}=\vect{k}_{\vect{f}*}^\top\boldsymbol{\alpha}$, where $\boldsymbol{\alpha}\in{\mathbb R}^N$ are the dual weights in feature space. 
The whole procedure only depends on a very small set of \blue{parameters}; collectively grouped in $\vect{\theta}$ for GP and in $\vect{\psi}$ for WGP. Inference of the \blue{parameters} and the weights $\boldsymbol{\alpha}$ can be performed using continuous optimization of the evidence in~\ref{eq:logevidence}.

For both the GP and WGP models we need to define the covariance (kernel, or Gram) function $k(\cdot,\cdot)$, which should capture the similarity between samples. We used the standard Automatic Relevance Determination (ARD) covariance $k(\vect{x}_i,\vect{x}_j) = \nu\exp(-\sum_{d=1}^D ({x}_i^d-{x}_j^d)^2/(2\sigma_d^2))$, which has given good performance in many problems~\cite{Rasmussen2006,CampsValls16grsm}, \blue{where $\vect{x}_i$ and $\vect{x}_j$ are vectors}. For GP, model \blue{parameters} are collectively grouped in $\vect{\theta}=\{\nu,\sigma_n,\sigma_1,\ldots,\sigma_d\}$. For the WGP we need to define a parametric smooth and monotonic form for $g$. In this paper we used:
$$g(y_i;\vect{\psi}) = \sum_{\ell=1}^L a_\ell~\text{tanh}(b_\ell~y_i + c_\ell),~~~~a_\ell,~b_\ell\geq 0, $$ 
where $\vect{\psi}=\{{\bf a},{\bf b},{\bf c}\}$. Even though any other sensible parametrization could be used, this one is quite convenient since it yields a set of {\em smooth steps} whose size, steepness and position are controlled by $a_\ell$, $b_\ell$ and $c_\ell$ parameters, respectively. \blue{In the present work, we fixed $L=5$ after several experiments, while the latter parameters were inferred via standard maximum log-likelihood maximization.}
%Recently, flexible non-parametric functions have replaced such parametric forms~\cite{Lazaro12warp}, thus placing another prior for $g(\x)\;\sim\;\GP(f,c(f,f'))$, whose model is learned via variational inference.

\begin{figure}[h!]
%\begin{center}
%\hspace{-1cm}
{\includegraphics[width=1\columnwidth]{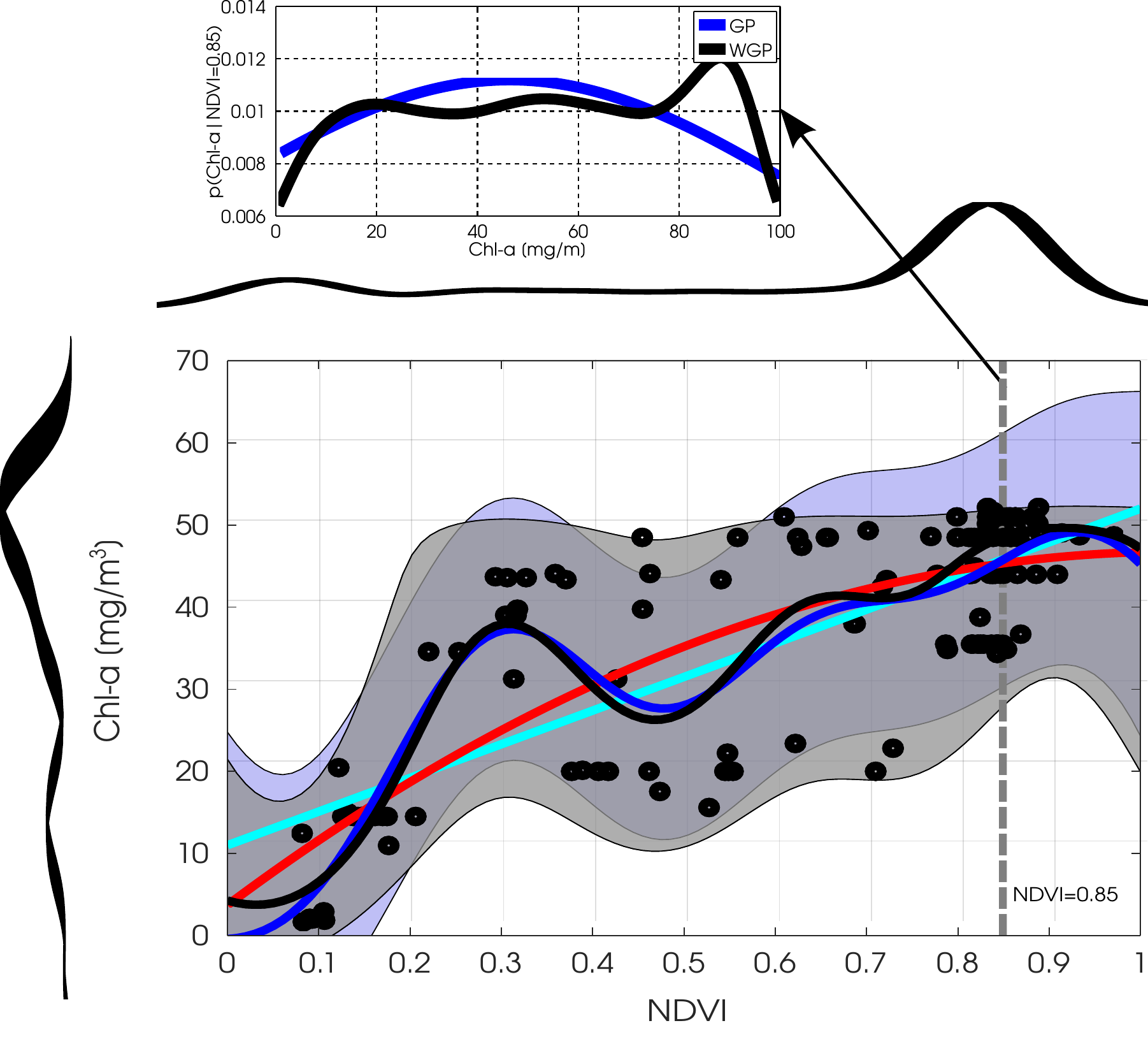}}
\caption{\blue{Scatterplot of NDVI vs LAI with predictive means (lines) and deviations (shaded areas) as well as the marginal distributions. The conditional densities at NDVI=0.85 (top of the figure) shows how the standard GP confidence intervals are symmetric while the WGP better captures the conditional distribution of the sample and yields asymmetric confidence intervals tighter around the more densely populated Chl-a regions.}\label{fig:warp2}}
%\end{center}
\end{figure}

\iffalse
\begin{figure}[t!]
%\hspace{3.4cm}
\centering
\includegraphics[width=.45\columnwidth]{./images/panel4.pdf}
\begin{center}
\vspace{-.7cm}
  \includegraphics[width=.95\columnwidth]{./images/panel1.pdf} \hspace{-1cm}\includegraphics[width=1cm,height=5.6cm]{./images/panel2.pdf}\\
\vspace{-.2cm}
\reflectbox{\rotatebox[origin=c]{180}{\includegraphics[width=8.4cm,height=1cm]{./images/panel3.pdf}}}
\end{center}
\caption{[Bottom] scatter of data with means (lines) and deviations (areas), and marginal distributions. Dashed line indicates NDVI level of 0.85 and in [Top] densities at NVDI=0.85 level of the standard GP (almost symmetric) and WGP (able to better capture the density of the sample). CREC QUE ES PODRIA COMPACTAR LA FIGURA; 1) LES MARGINALS EN ALTRES A ESQUERRA I DALT, LA CONDICIONAL A LA DRETA EN UNA FLETXETA, I ARMONITZAR ELS SEUS ESTILS, QUEDA UN PASTITXE. 2) LES CONFIANCES ESTAN PER DAMUNT DE LES MITJANES. } \label{fig:warp2}}
\end{figure}
\fi

%%%%%%%%%%%%%%%%%%%%%%%%%%%%%%%%%%%%
\subsection{Illustration of a warped GP model}

%\blue{FROM SHSIC PAPER: This section illustrates and discusses the behavior of both standard GP and WGP in a real problem. In particular, we deal with a regression setting, trying to link satellite observations with {\em in-situ} measurements of chlorophyll content (Chl-a) of plants in the land cover. The data used in this study were obtained in two terrestrial campaigns in Barrax, Spain. The test area has a rectangular form and an extent of 5 km $\times$ 10 km, and is characterized by a flat morphology and large, uniform land-use units. The region consists of approximately 65\% dry land and 35\% irrigated land. We used a calibrated CCM-200 Chlorophyll Content Meter to estimate Chl-a. Simultaneously, we used satellite images from the Compact High Resolution Imaging Spectrometer (CHRIS) sensor, which is the prime instrument of the PROBA-1 mission. The technology objective is to explore the capabilities of imaging spectrometers on agile small satellite platforms. CHRIS measures over the visible/ near-infrared spectra from 400 to 1050 nm. For this study, we used CHRIS data in Mode 1 (62 bands, full spectral information) for the four campaign days, where {\em in situ} measurements of surface properties were measured in conjunction with the satellite overpass. The images were geometrically and atmospherically corrected. A total set of $135$ measurements in the 62-dimensional feature space constitute the database. Further details can be found in~\cite{Verrelst12}.}

We illustrate the main difference between GPs and WGPs in the problem of estimating chlorophyll concentration (Chl-a) from the NDVI using {\em in situ} measurements (data description and experiments further explained in section~\ref{sec:experiment2}). Figure~\ref{fig:warp2} shows the data distribution, as well as linear (cyan), second order polynomial (red), GP (blue), and WGP (black) fitting. In gray and blue shaded areas we provide the confidence intervals (mean$\pm$2 standard deviations) for the GP and WGP models, respectively. The marginals indicate a denser sampling of Chl-a in the NDVI region between [0.7,0.9], and reveal a non-Gaussian distribution of the Chl-a target variable. 
%an scatterplot of the NDVI vs. Chlorophyll acquired measurements. The bottom part of the figure illustrates different regression algorithms models: (1) standard linear regression, (2) quadratic fit, (3) GP, and (4) WGP. Continuous lines indicates the mean of the models and the areas (overlapped somewhere) are the confidence intervals of the GP's. 
The main difference between standard GP and WGP arises on the shape of its confidence intervals, which is better observed in the conditional pdf, i.e. $p(y|\text{NVDI}=0.85)$. \blue{In top of Fig. \ref{fig:warp2}, the standard GP yields a symmetric area around its mean, but the shape of the predictive distribution of WGP is more flexible and adapts better to the data distribution. Note that the conditional confidence interval in this case shows three different peaks, revealing higher confidence around the chlorophyll regions more densely sampled.} 
%dominant modes across the Chl-a range.} %We also shown densities of GP models given a particular value of NDVI=0.85, \blue{$(p(y|NVDI=0.85))$} and as it can be seen in the plot located at the top of the figure, WGP is able to find and characterize properly the density distribution by obtaining 3 different peaks which represents regions of the space where more points are concentrated dealing to more density.

\iffalse
\begin{figure}[t!]
%\centerline{\includegraphics[width=7cm]{./images/panel1.pdf}} &
%\centerline{\includegraphics[width=7cm,height=1cm]{./images/panel2.pdf}} \\
\includegraphics[width=8cm]{./images/scatterhist.png} \\
\includegraphics[width=8cm,height=5cm]{./images/panel4.pdf}  \\
\caption{\blue{Adrian!!!}. \label{fig:warp2}}
\end{figure}
\begin{figure}[t!]
\centering
  \includegraphics[width=.8\columnwidth]{./images/panel1.pdf} \hspace{-1cm}\includegraphics[width=.15\columnwidth,height=4.7cm]{./images/panel2.pdf}\\
\vspace{-.7cm}
\includegraphics[width=7cm,height=1cm,angle=180]{./images/panel3.pdf} 
\includegraphics[width=.7\columnwidth]{./images/panel4.pdf}
\caption{\blue{Adrian!!!}. \label{fig:warp2}}
\end{figure}
\fi

%%%%%%%%%%%%%%%%%%%%%%%%%%%%%%%%%%%%
%%%%%%%%%%%%%%%%%%%%%%%%%%%%%%%%%%%%
%%%%%%%%%%%%%%%%%%%%%%%%%%%%%%%%%%%%
%%%%%%%%%%%%%%%%%%%%%%%%%%%%%%%%%%%%
%%%%%%%%%%%%%%%%%%%%%%%%%%%%%%%%%%%%

\section{Experimental results}\label{sec:exp}

We show the potential of WGP in the following three experiments:
1) estimation of ocean chlorophyll concentration from multispectral data;
2) vegetation parameter retrieval from hyperspectral images; and
3) causal inference in a set of 28 geosciences problems.

%%%%%%%%%%%%%%%%%%%%%%%%%%%%%%%%%%%%
\subsection{Experiment 1: Ocean chlorophyll estimation}\label{sec:experiment1}

%\paragraph{Data collection}\label{sec:data}

%%We focus here on the estimation of chlorophyll-a concentrations from remote sensing upwelling radiance just above the ocean surface. A variety of bio-optical algorithms have been developed to relate measurements of ocean radiance to {\em in situ} concentrations of phytoplankton pigments, and ultimately most of these algorithms demonstrate the potential of quantifying chlorophyll-a concentrations accurately from multispectral satellite ocean color data. \blue{In this context, robust and stable non-linear regression methods that provide inverse models are desirable. In addition, we should note that most of the bio-optical models (such as Morel, CalCOFI and OC2/OC4 models) often rely on empirically adjusted nonlinear transformation of the observed variable (which is traditionally a ratio between bands).}

We focus here on the estimation of chlorophyll-a concentrations from remote sensing upwelling radiance just above the ocean surface. A variety of bio-optical algorithms have been developed to relate measurements of ocean radiance to {\em in situ} concentrations of phytoplankton pigments, and ultimately most of these algorithms demonstrate the potential of quantifying chlorophyll-a concentrations accurately from multispectral satellite ocean color data. In addition, we should note that most of the bio-optical models (such as Morel, CalCOFI and OC2/OC4 models) often rely on empirically adjusted nonlinear transformation of the observed variable (which is traditionally a ratio between bands). In this context, more robust and stable non-linear regression methods are desirable.

We used the SeaBAM dataset~\cite{oreilly98}, %,maritorena00}
which gathers 919 {\em in situ} pigment measurements around the United States and Europe. The dataset contains coincident {\em in situ} chlorophyll concentration and remote sensing reflectance measurements (Rrs($\lambda$), [sr$^{-1}$]) at some wavelengths (412, 443, 490, 510 and 555 nm) that are present in the SeaWiFS ocean color satellite sensor. The chlorophyll concentration values range varies from 0.019 to 32.79 mg/m$^3$, revealing a clear exponential distribution. %), mainly due to the fact that SeaBAM data is originated from various teams and campaigns. 
%%%At high Chl-a  concentrations, Ca [mg/m$^3$], the dispersion of radiance ratios  Rrs(490)/Rrs(555) increases, mostly because of the presence of Case II waters. The shape of the scatterplots is approximately sigmoidal in log-log space. At lowest concentrations the highest Rrs(490)/Rrs(555) ratios are slightly lower than the theoretical limit for clear natural waters. See analysis in~\cite{Camps-Valls2006}. 
More information about the data can be obtained from \url{http://seabass.gsfc.nasa.gov/seabam/seabam.html}.

\begin{table*}[t!]
\begin{center}
\caption{Bias (ME), accuracy (RMSE, MAE) and fitness (\blue{coefficient of correlation} R) for different rates of training samples using both raw and empirically-transformed variable for the SeaBAM dataset.}\label{res}
\renewcommand{\tabcolsep}{4pt}
\begin{tabular}{|c|c|c|c|c||c|c|c|c|}
\hline
\hline
& \multicolumn{4}{|c||}{Raw} & \multicolumn{4}{c|}{Empirical}\\ 
\hline
\hline
	 & ME	 & RMSE	 & MAE	 & R 	 & ME	 & RMSE	 & MAE	 & R \\ 
\hline
%\hline
%\multicolumn{9}{|l|}{rate = 5\%}\\ 
%\hline
%GP	 & 0.29	 & 2.31	 & 0.45	 & 0.62 & 0.27 & 2.30 & 0.44 & 0.63 \\ 
%VHGP & 0.29	 & 2.36	 & 0.48	 & 0.59 & 0.27 & 2.29 & 0.44 & 0.63 \\ 
%WGP	 & 0.29	 & 2.33	 & 0.45	 & 0.65 & 0.27 & 2.20 & 0.43 & 0.65 \\ 
%\hline
%\multicolumn{9}{|l|}{rate = 10\%}\\ 
%\hline
%GP	 & 0.38	 & 2.54	 & 0.51	 & 0.52 & 0.26	 & 2.12	 & 0.42	 & 0.71 \\ 
%VHGP & 0.37	 & 2.49	 & 0.49	 & 0.56 & 0.26	 & 2.12	 & 0.42	 & 0.71 \\ 
%WGP	 & 0.30	 & 2.32	 & 0.45	 & 0.67 & 0.31	 & 2.36	 & 0.45	 & 0.61 \\ 
%\hline
%\multicolumn{9}{|l|}{rate = 20\%}\\ 
%\hline
%GP	 & 0.30	 & 2.31	 & 0.45	 & 0.74 & 0.33	 & 2.33	 & 0.42	 & 0.75 \\ 
%VHGP & 0.43	 & 2.62	 & 0.51	 & 0.68 & 0.33	 & 2.36	 & 0.43	 & 0.74 \\ 
%WGP	 & 0.26	 & 2.20	 & 0.40	 & 0.77 & 0.34	 & 2.34	 & 0.43	 & 0.76 \\ 
%\hline
%\multicolumn{9}{|l|}{rate = 40\%}\\ 
%\hline
%GP	 & 0.02	 & 1.74	 & 0.33	 & 0.82 & 0.15	 & 1.69	 & 0.29	 & 0.86 \\ 
%VHGP & 0.29	 & 2.51	 & 0.46	 & 0.65 & 0.15	 & 1.70	 & 0.29	 & 0.85 \\  
%WGP	 & 0.08	 & 1.71	 & 0.30	 & 0.83 & 0.17	 & 1.75	 & 0.30	 & 0.86 \\ 
\hline
\multicolumn{9}{|l|}{\blue{rate = 20\%}}\\ 
\hline
GP &0.146 $\pm$ 0.142&1.836 $\pm$0.541&0.490 $\pm$0.175&0.765 $\pm$0.139&0.148 $\pm$ 0.112&2.390 $\pm$1.001&0.411 $\pm$0.098&0.736 $\pm$0.132\\
VHGP &0.252 $\pm$ 0.129&2.195 $\pm$0.342&0.530 $\pm$0.147&0.657 $\pm$0.119&0.141 $\pm$ 0.105&2.384 $\pm$1.017&0.410 $\pm$0.096&0.737 $\pm$0.134\\
WGP &0.143 $\pm$ 0.110&1.710 $\pm$0.337&0.401 $\pm$0.055&0.804 $\pm$0.071&0.146 $\pm$ 0.118&2.583 $\pm$1.122&0.445 $\pm$0.116&0.686 $\pm$0.149\\
\hline
\multicolumn{9}{|l|}{\blue{rate = 50\%}}\\ 
\hline
GP &0.133 $\pm$ 0.110&2.036 $\pm$0.468&0.453 $\pm$0.057&0.765 $\pm$0.106&0.149 $\pm$ 0.144&2.024 $\pm$1.689&0.349 $\pm$0.142&0.823 $\pm$0.130\\
VHGP &0.183 $\pm$ 0.119&1.786 $\pm$0.454&0.398 $\pm$0.046&0.789 $\pm$0.090&0.146 $\pm$ 0.116&1.934 $\pm$1.272&0.343 $\pm$0.115&0.828 $\pm$0.117\\
WGP &0.069 $\pm$ 0.050&1.318 $\pm$0.331&0.309 $\pm$0.042&0.881 $\pm$0.054&0.107 $\pm$ 0.065&1.607 $\pm$0.340&0.326 $\pm$0.051&0.832 $\pm$0.064\\
\hline
\multicolumn{9}{|l|}{\blue{rate = 80\%}}\\ 
\hline
GP &0.122 $\pm$ 0.180&1.620 $\pm$0.848&0.457 $\pm$0.160&0.836 $\pm$0.148&0.155 $\pm$ 0.138&1.477 $\pm$0.882&0.333 $\pm$0.155&0.889 $\pm$0.089\\
VHGP &0.211 $\pm$ 0.192&1.873 $\pm$0.787&0.430 $\pm$0.158&0.811 $\pm$0.113&0.132 $\pm$ 0.114&1.306 $\pm$0.699&0.306 $\pm$0.119&0.906 $\pm$0.075\\
WGP &0.113 $\pm$ 0.072&1.272 $\pm$0.671&0.310 $\pm$0.099&0.885 $\pm$0.129&0.146 $\pm$ 0.139&1.536 $\pm$0.798&0.344 $\pm$0.140&0.889 $\pm$0.068\\
\hline
\hline
\end{tabular}
\end{center}
\end{table*}

{Table~\ref{res} shows different scores: mean error (ME), accuracy (RMSE \& MAE), coefficient of \blue{correlation (R)} between the observed and predicted variable when using the raw data (no {\em ad hoc} transform at all) and the} empirically-adjusted transform\footnote{Several transformations were tested: log, exp, and polynomial.}. %We did two experiments: (1) working with raw data and (2) working with an empirical transformed variable. 
Results are shown for three flavours of GPs: the standard GP regression (GP)~\cite{Rasmussen2006}, the variational heteroscedastic GP (VHGP)~\cite{CampsVallsGRSL2013}, and the proposed warped GP regression (WGP)~\cite{Snelson03} for different rates of training samples. Several conclusions can be obtained:
1) better results are obtained by all models when using more training samples,
2) \blue{for GP and VHGP, using empirically-based features improves the results over using raw data for the same number of training samples},  %, especially for high rates}, %but this requires prior knowledge about the problem, time and efforts to fit an appropriate function, 
3) WGP outperforms standard GP and VHGP in all comparisons \blue{when using raw data}, therefore
4) \blue{results show that WGP compensates the lack of prior knowledge about the (skewed) distribution of the variable}.

\begin{figure}[b!]
\vspace{-0.5cm}
\centerline{\includegraphics[width=7cm]{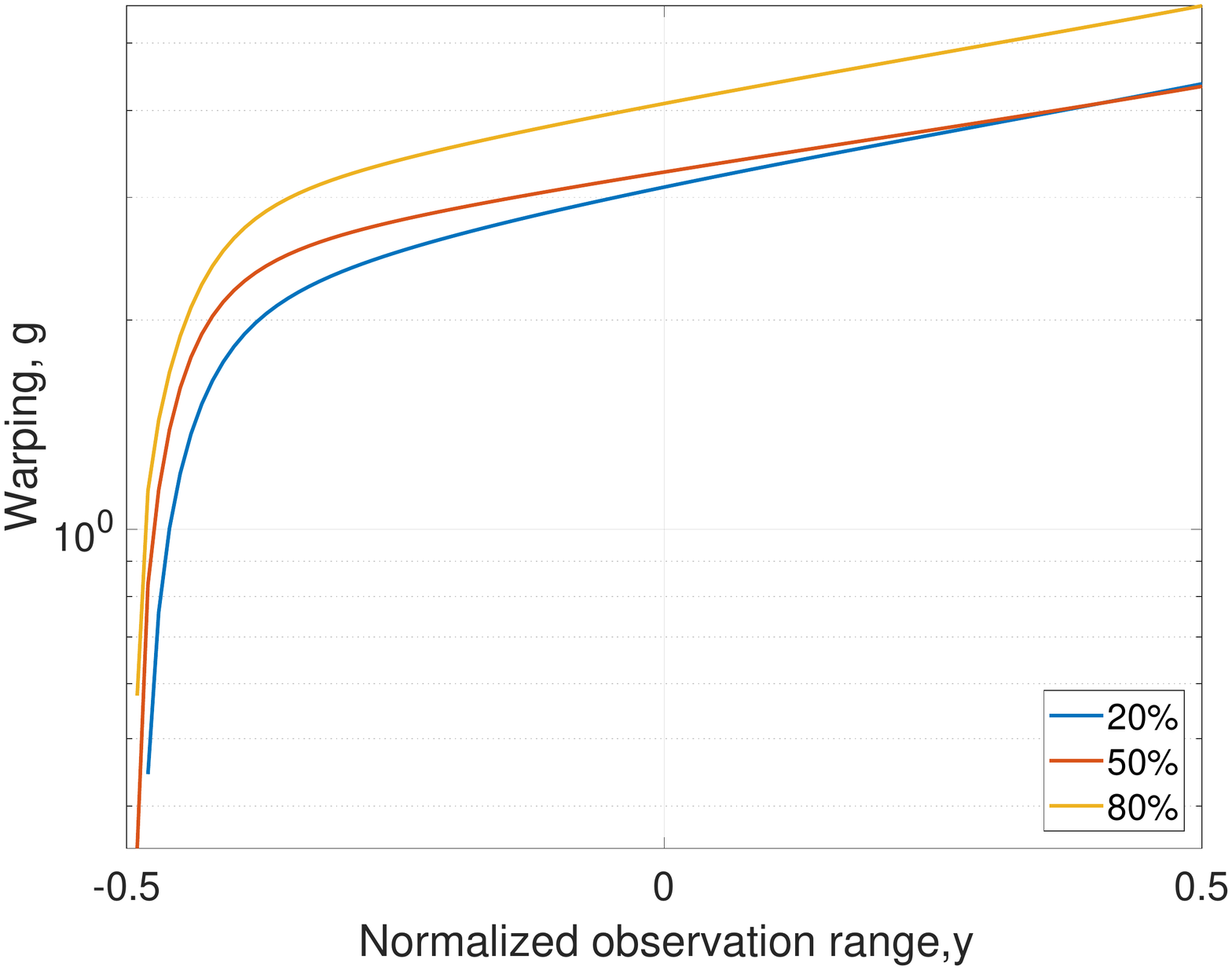}}
\vspace{-0.25cm}
\caption{Learned warping transformations with WGP for different rates of training samples. The range of $y$ is normalized $[-1,1]$ for better representation.}
\label{fig:warp}
%\end{wrapfigure}
\end{figure}

An interesting advantage of WGP is we have access to the learned warping function. We plot these for different rates of training samples in Fig.~\ref{fig:warp}, which shows that:
1) as more samples are provided for learning, the warping function becomes more nonlinear for low chlorophyll concentration values;
2) the learned warping function actually looks linear (in log-scale) for high observation values and strongly nonlinear for low values. The empirically-based warping function typically used in most bio-optical models is a log function. Therefore it seems that the WGP accuracy comes from the better modeling of the nonlinearity for low chlorophyll values, which are the great majority in the database.

%\begin{wrapfigure}{r}{5cm}

%\begin{figure}[b!]
%\vspace{-0.5cm}
%\centerline{\includegraphics[width=7cm]{./images/warping_solution.pdf}}
%\vspace{-0.25cm}
%\caption{Learned warping transformations with WGP for different rates of training samples. The range of $y$ is normalized $[-1,1]$ for better representation.}
%\label{fig:warp}
%\end{wrapfigure}
%\end{figure}

%%%%%%%%%%%%%%%%%%%%%%%%%%%%%%%%%%%%
\subsection{Experiment 2: Estimation of vegetation parameters}\label{sec:experiment2}

%\blue{COMPLETAR ELS EXPERIMENTS. 1) NO SEMBLA QUE WGP GUANYE (SI NO RESULTA EVIDENT PODRIEU UTILITZAR UNA ALTRA BD); 2) FALTA VHGP I ELIMINAR RF ; 3) LA FIGURA 3  SENSE EIXOS, UN UNIC COLORBAR,}

In this second experiment, we used data from the SPARC campaign acquired during the month of July of 2003 in Barrax (Spain), previously mentioned in Sec. \ref{sec:wgp}. We have \textit{in-situ} ground-truth data of three different \blue{vegetation parameters}, chlorophyll-a (Chl-a), Leaf Area Index (LAI) and Fraction of Vegetation Cover (FVC). We trained the models using hyperspectral reflectances from the CHRIS instrument on board of PROBA-1 satellite\footnote{\url{https://earth.esa.int/web/guest/missions/esa-operational-eo-missions/proba}}, which provides information in 62 spectral channels at a spatial resolution of about 34 meters. 

%\blue{As in the preceding experiment, we tested three different methods, but now we switched VHGP method to Random Forest method (RF). The three tests used the same samples division for carry out the training. Thus, the three methods are in the same conditions for comparing them.}

As the number of training samples are scarce (only $n=135$ labeled samples in total), we report results using a $4$-fold cross validation procedure.
% The training set is divided into four folds, then we made three experiments, taking 3/4 for training and 1/4 for testing in each one. 
The results in Table~\ref{tab:res2} show the mean and the standard deviation for the experiments. Results are reported in the same terms that in the previous experiment~\ref{sec:experiment1}.
% in terms of mean error (ME), root mean squared error (RMSE), and squared Pearson's correlation coefficient (R$^2$).
% \blue{Table \ref{res2} shows different scores (mean error, root-mean-square error and  R$^2$) with their respective standard deviation after applying the three different methods previously mentioned. } 
The proposed warped GP regression (WGP) obtains \blue{better results for prediction of Chl-a content and FVC}. WGP achieves better quality estimates and lower error bars compared to the other GP models.

\begin{table}[h!] 
\begin{center}
\caption{Bias (ME), accuracy (RMSE) and fitness (R$^2$) for the tested vegetation variables (Chl-a, LAI, FVC) in the SPARC dataset.\label{tab:res2}}
\renewcommand{\tabcolsep}{2pt}
\begin{tabular}{|c|c|c|c|} \hline \hline
	 & ME	 & RMSE	 &  R$^2$ \\  \hline \hline
Chl-a &&& \\ \hline
GP	  & -0.99 $\pm$ 2.14  & 5.75 $\pm$ 3.81   & 0.845 $\pm$ 0.16 \\
VHGP  & -1.05 $\pm$ 2.15  & 5.58 $\pm$ 4.12   & 0.854 $\pm$ 0.17 \\
WGP   & -0.51 $\pm$ 1.89  & 5.16 $\pm$ 3.13   & 0.873 $\pm$ 0.12 \\
% RF   & -0.23 $\pm$ 2.32  & 6.25	$\pm$ 1.84   & 0.823 $\pm$ 0.06 \\ 
\hline
LAI &&& \\ \hline
GP	  & 0.005 $\pm$ 0.10  & 0.53 $\pm$ 0.07   & 0.879 $\pm$ 0.03 \\
VHGP  & 0.03 $\pm$ 0.15   & 0.55 $\pm$ 0.05   & 0.875 $\pm$ 0.03 \\
WGP   & 0.02 $\pm$ 0.14	  & 0.55 $\pm$ 0.11   & 0.867 $\pm$ 0.05 \\ 
% RF   & 0.04 $\pm$ 0.12   & 0.69 $\pm$ 0.14   & 0.792 $\pm$ 0.09 \\ 
\hline
FVC &&& \\ \hline
GP	  & 0.012 $\pm$ 0.009    & 0.142 $\pm$ 0.040   & 0.796 $\pm$ 0.105\\
VHGP  & -0.021 $\pm$ 0.013    & 0.144 $\pm$ 0.041   & 0.799 $\pm$ 0.088\\
WGP   & 0.006 $\pm$ 0.007   & 0.137 $\pm$ 0.038   & 0.814 $\pm$ 0.090\\ 
% RF   & 0.014 $\pm$ 0.02   & 0.13 $\pm$ 0.02   & 0.83 $\pm$ 0.08\\ 
\hline \hline
\end{tabular}
\end{center}
\end{table}

\begin{figure}[t!]
\begin{center}
\setlength{\tabcolsep}{-1mm}
%\begin{tabular}{ p{2.5cm} p{2.5cm} p{2.5cm} }
\begin{tabular}{ccc}
GP &  VHGP & WGP \\
\includegraphics[scale=0.1]{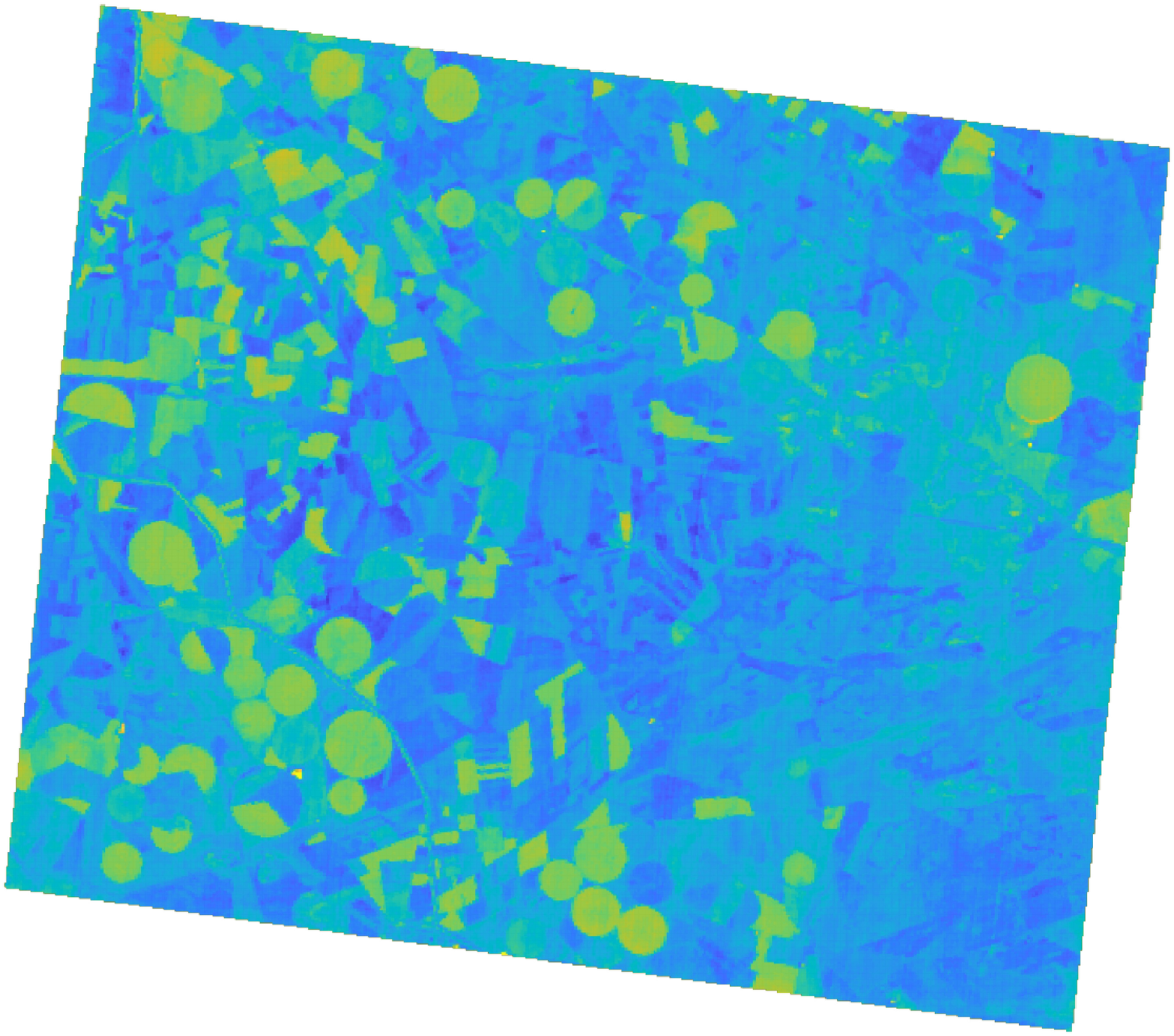} & 
\includegraphics[scale= 0.1]{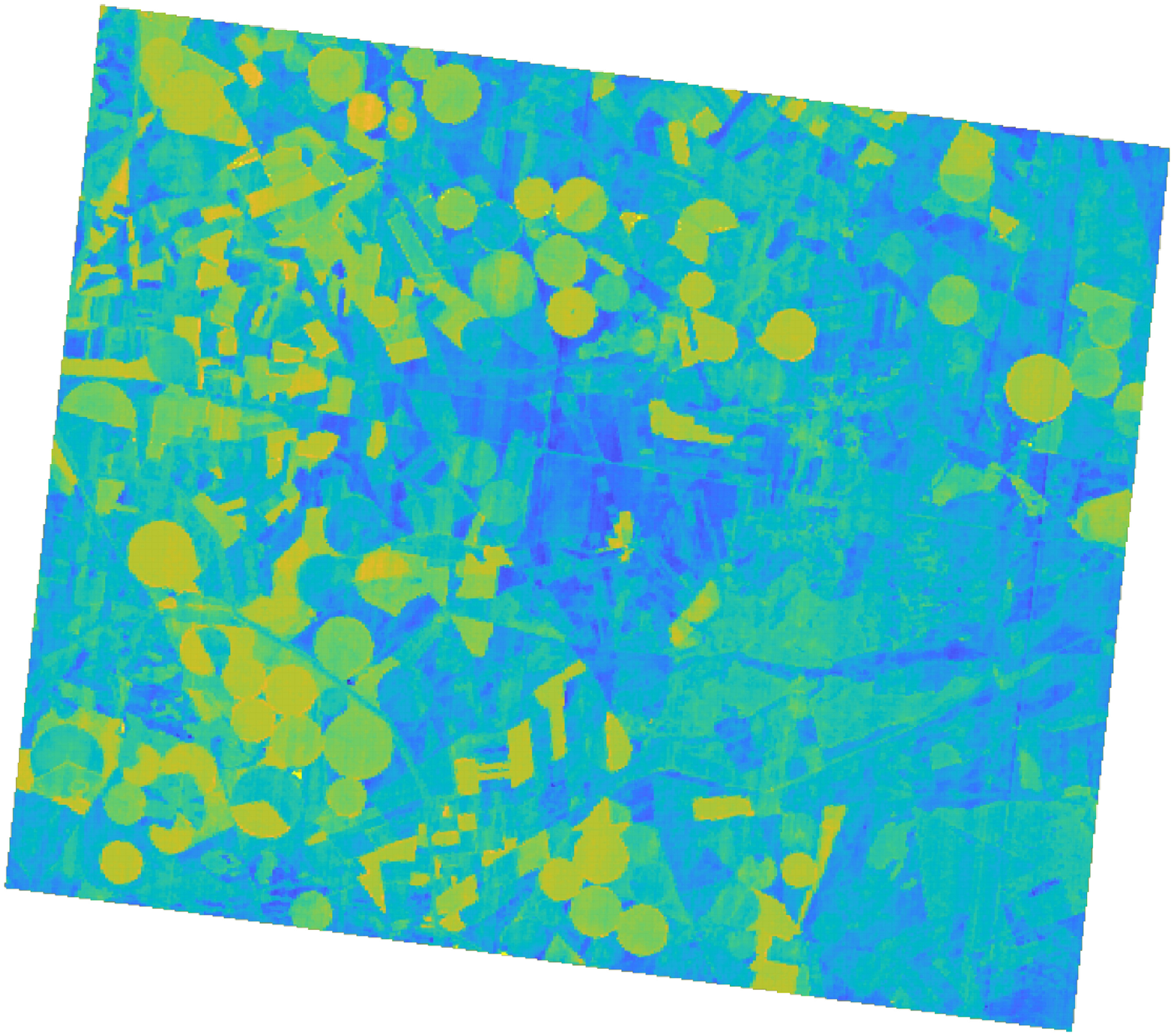}& 
\includegraphics[scale= 0.1]{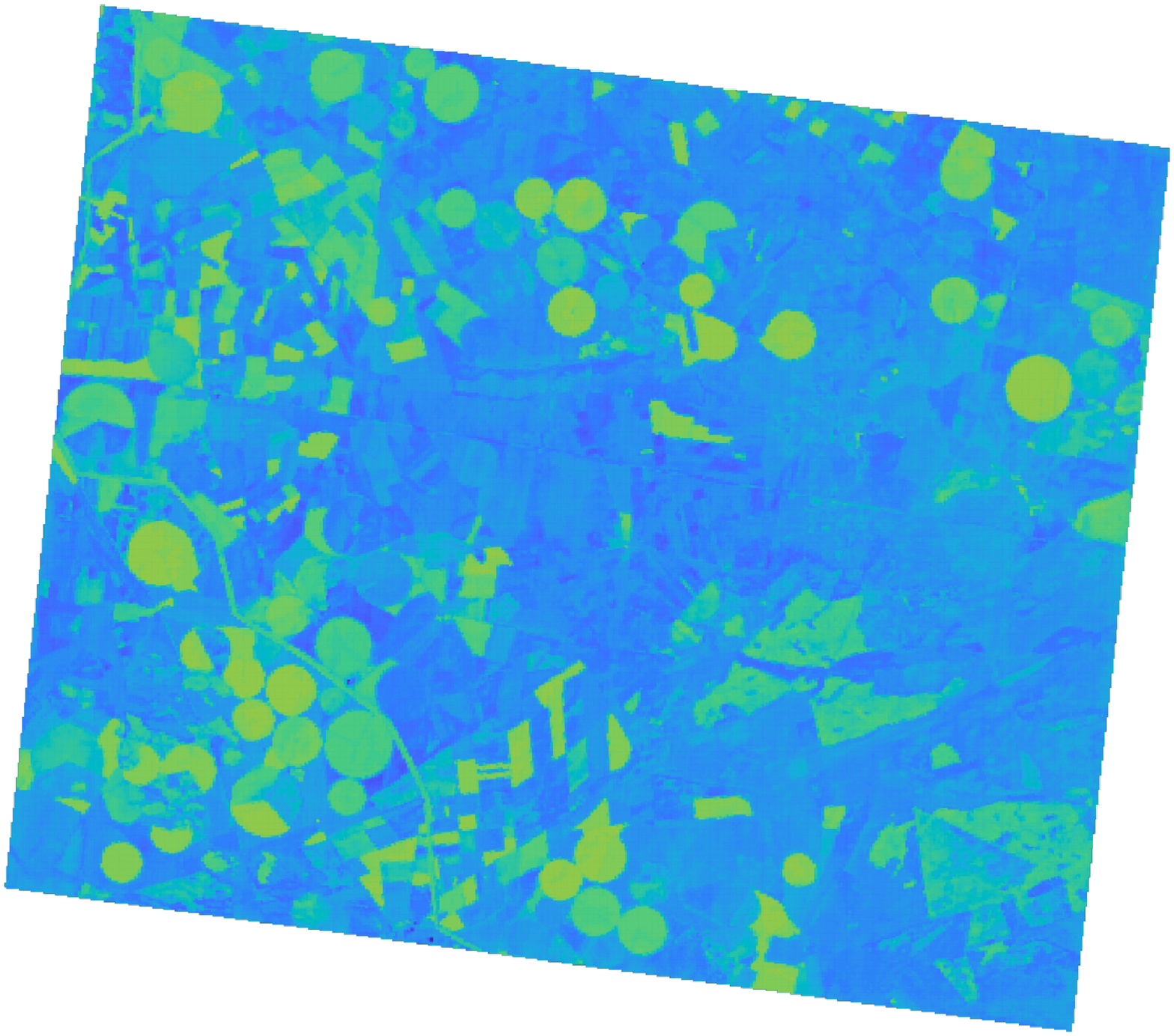} 
\includegraphics[scale= 0.025]{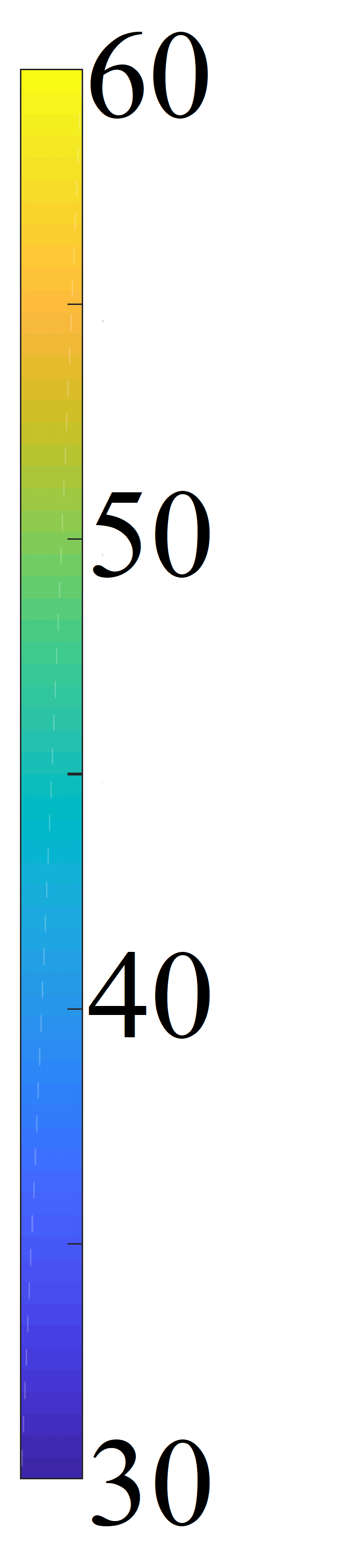} \\
\includegraphics[scale= 0.1]{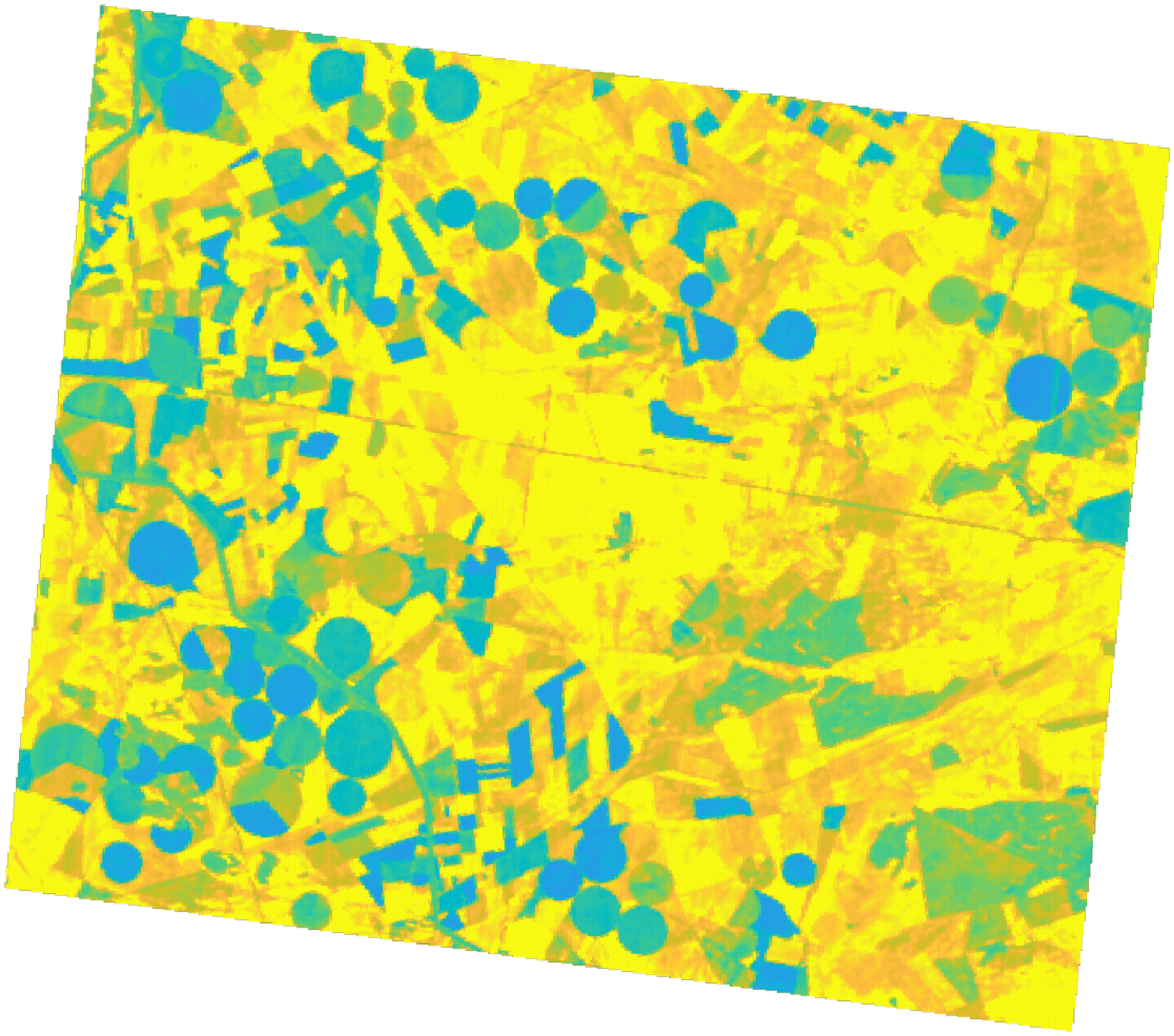} & 
\includegraphics[scale= 0.1]{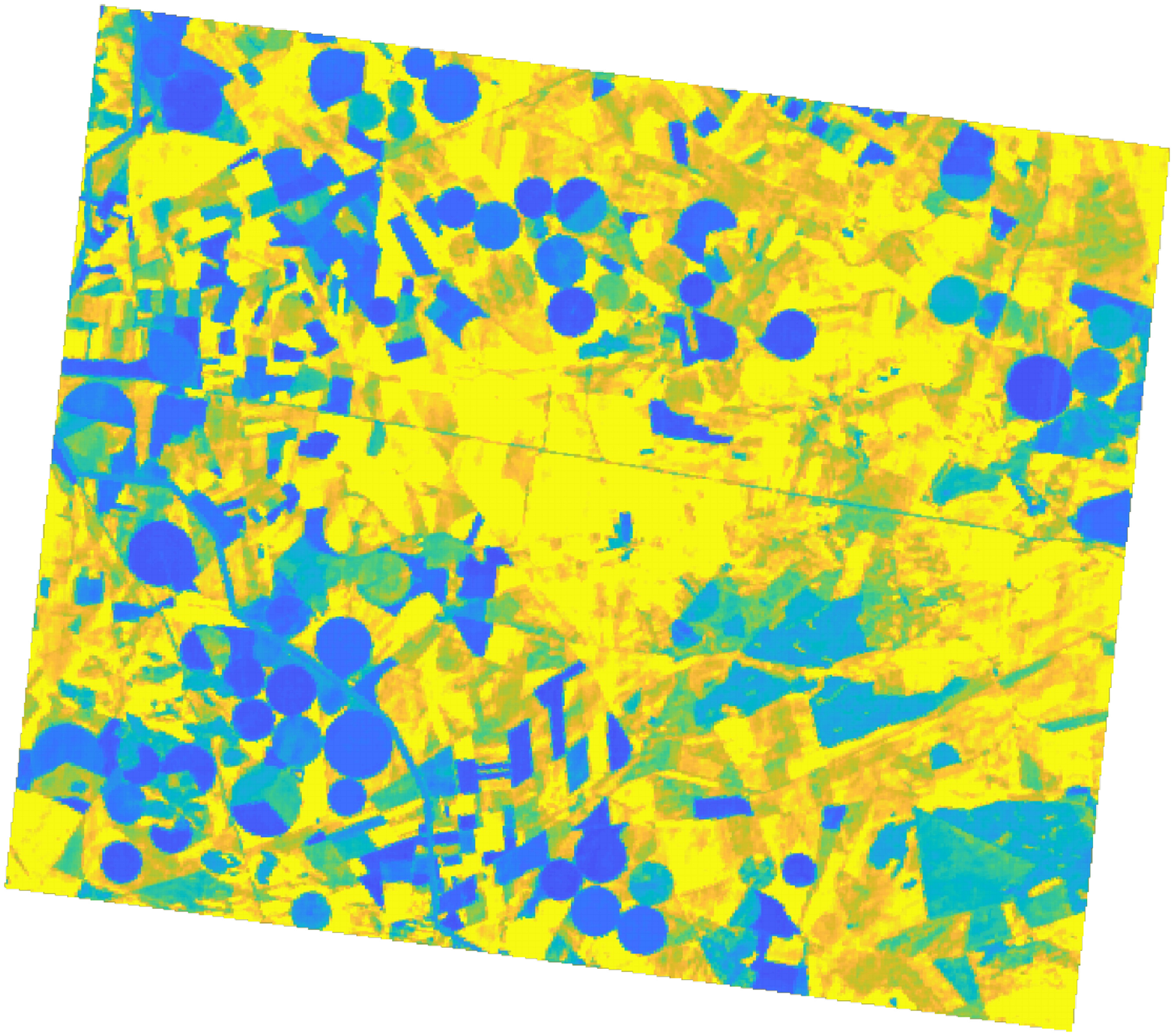}& 
\includegraphics[scale= 0.1]{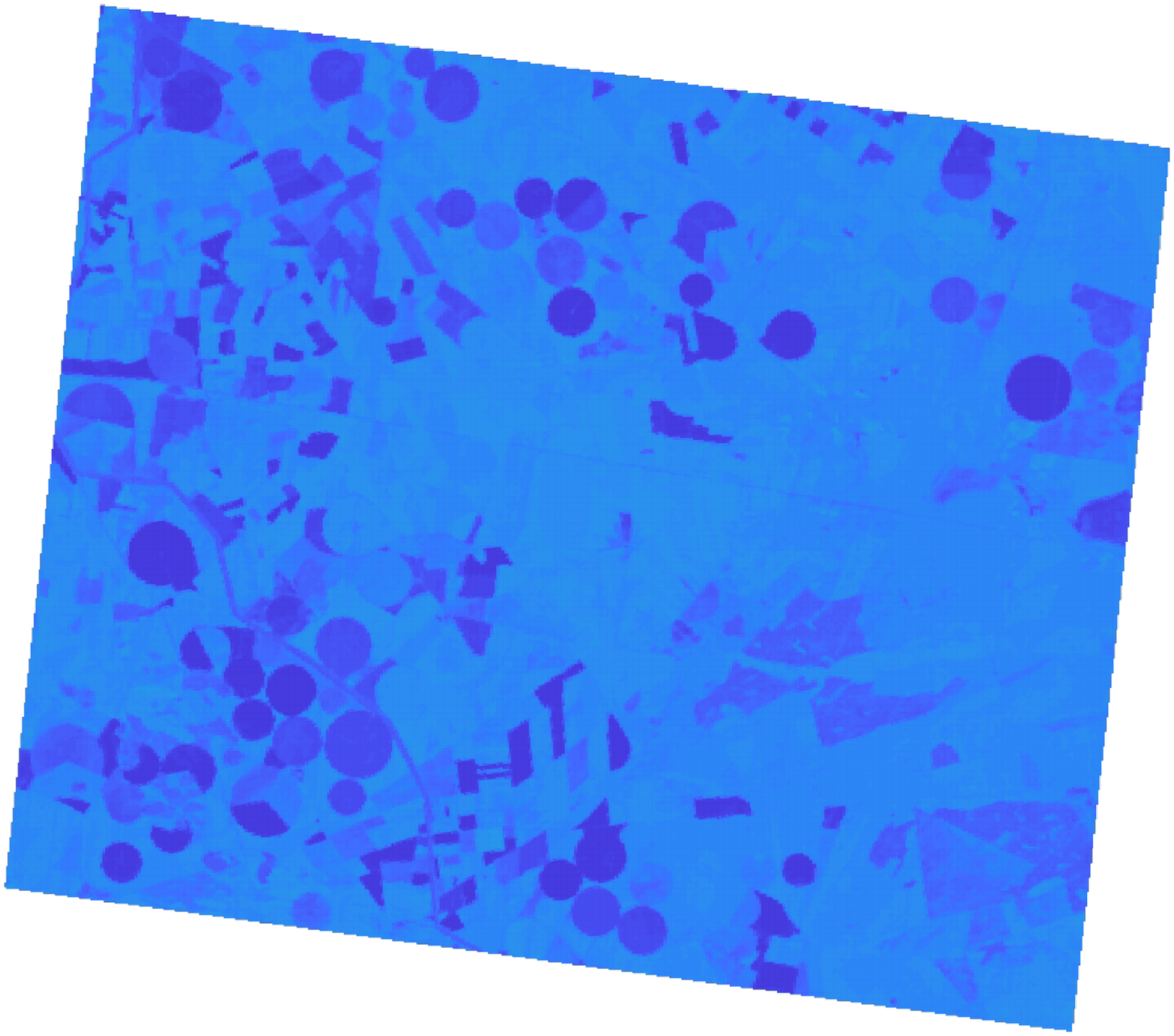}
\includegraphics[scale= 0.025]{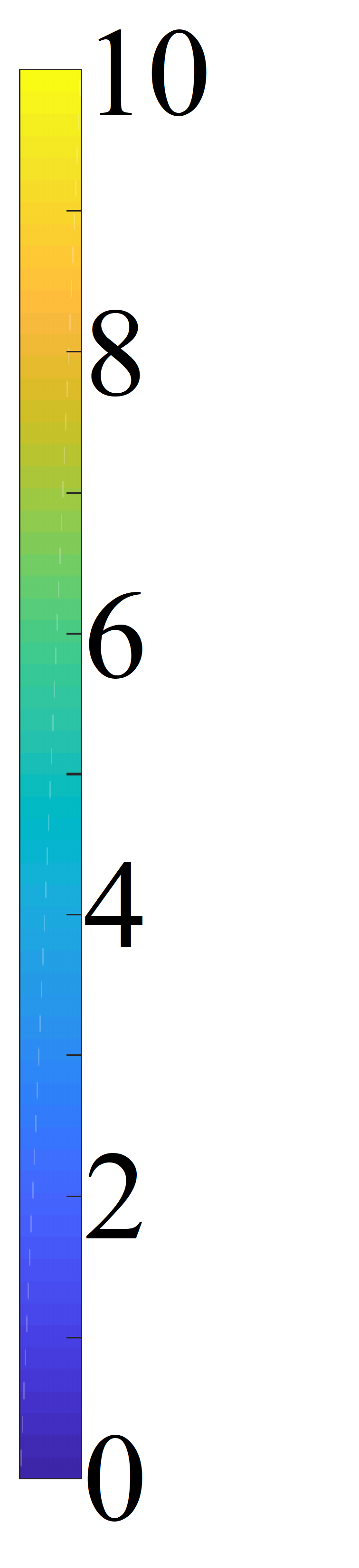} \\
\includegraphics[scale= 0.1]{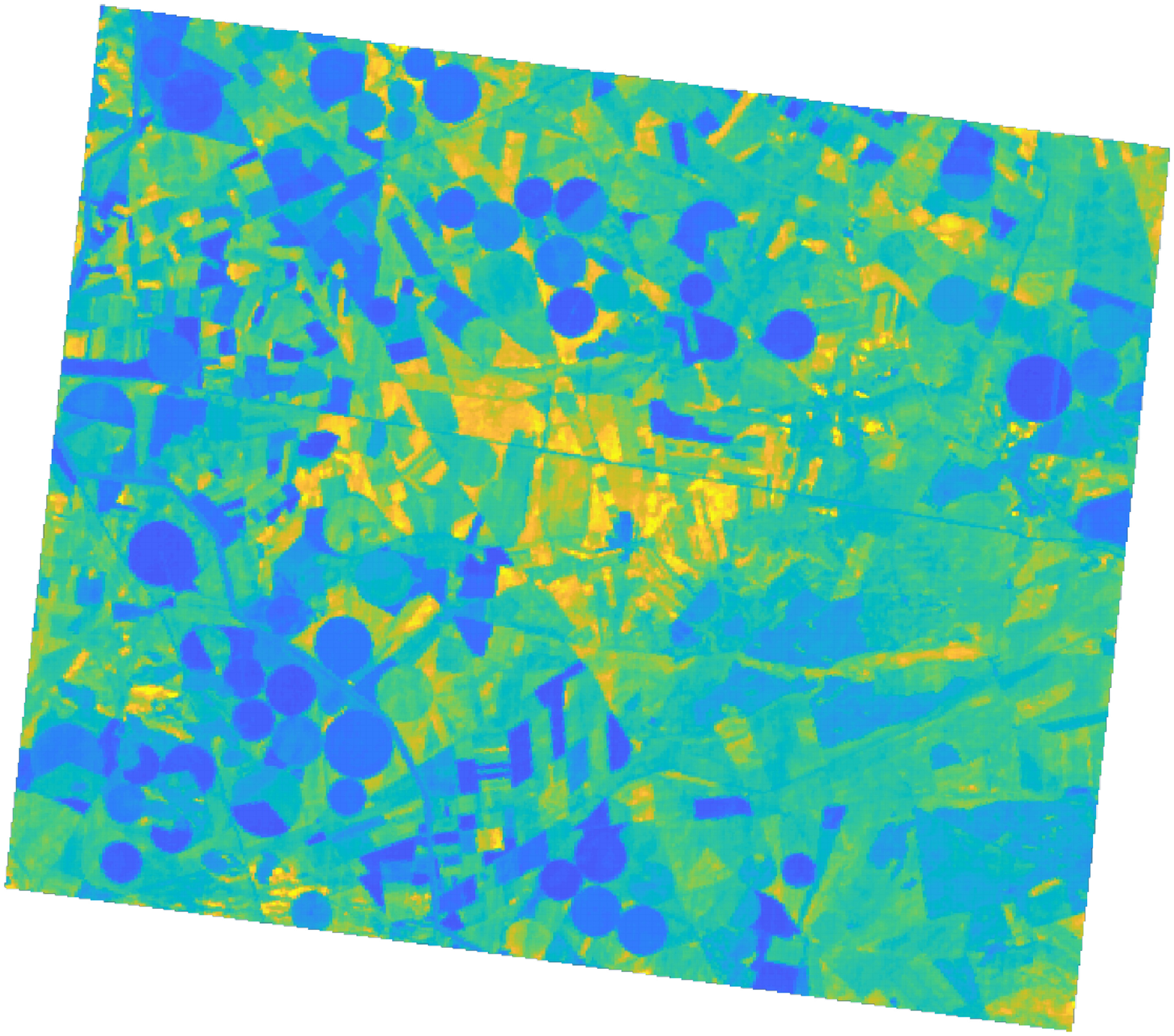} & 
\includegraphics[scale= 0.1]{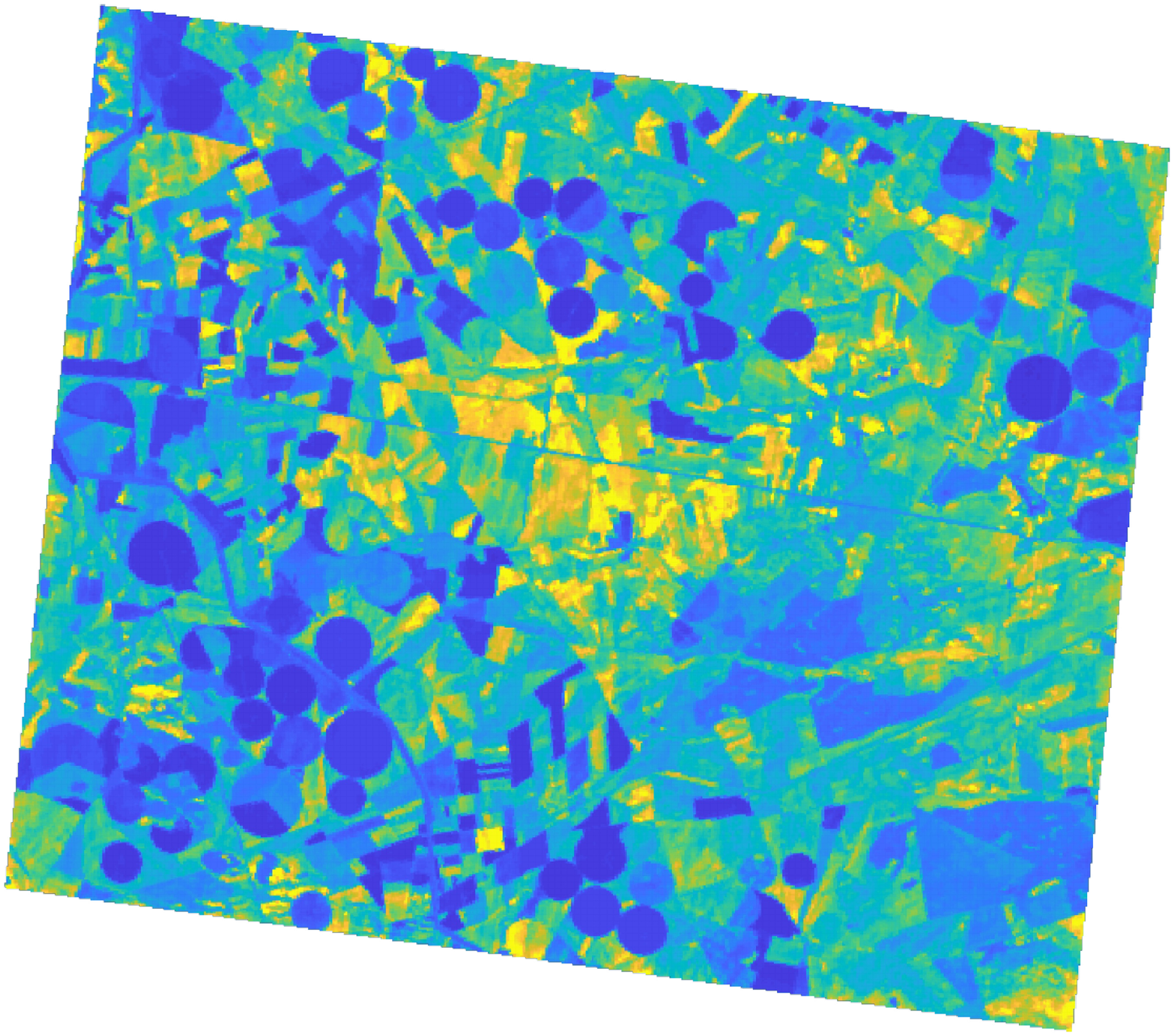}& 
\includegraphics[scale= 0.1]{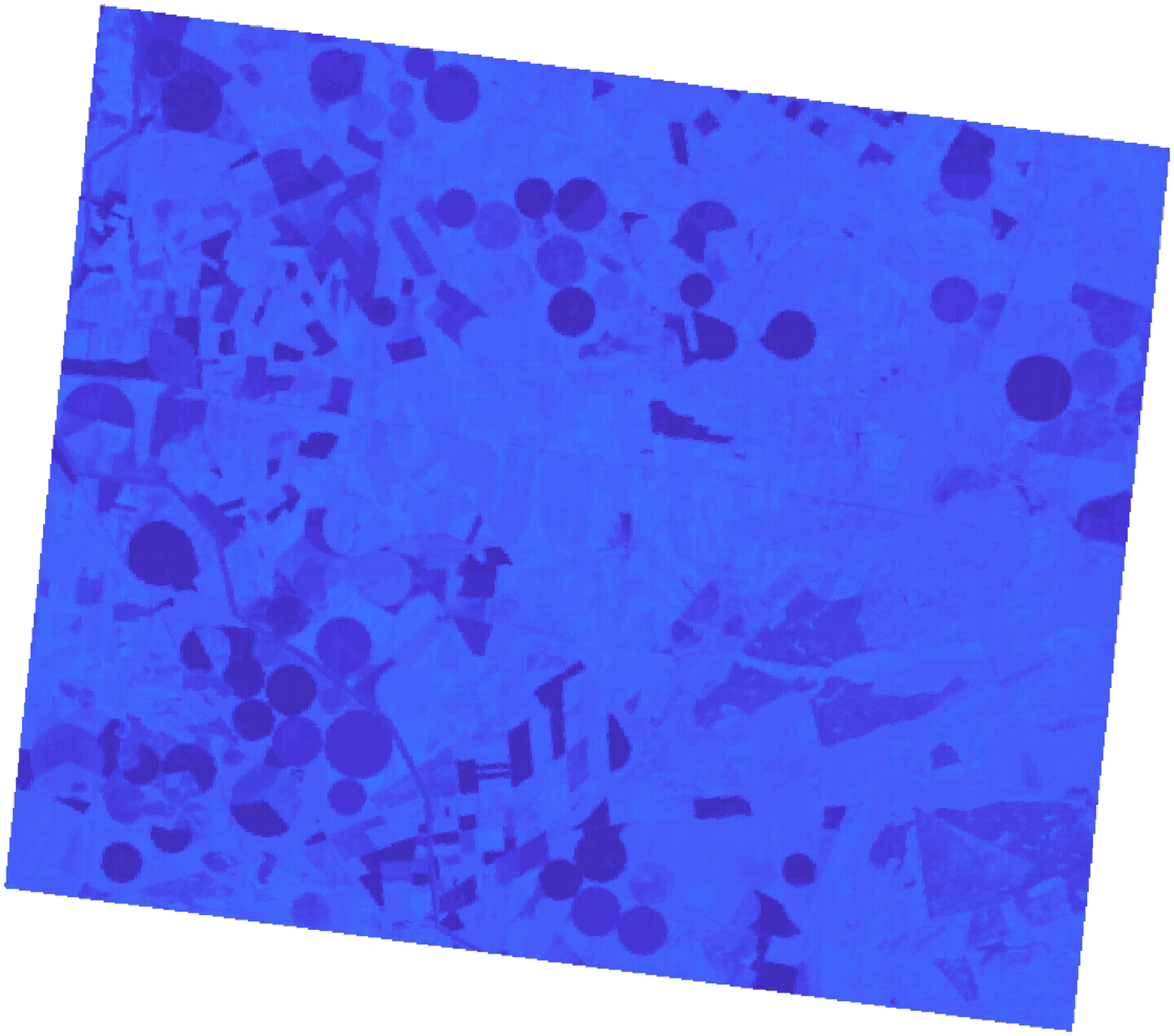}
\includegraphics[scale= 0.025]{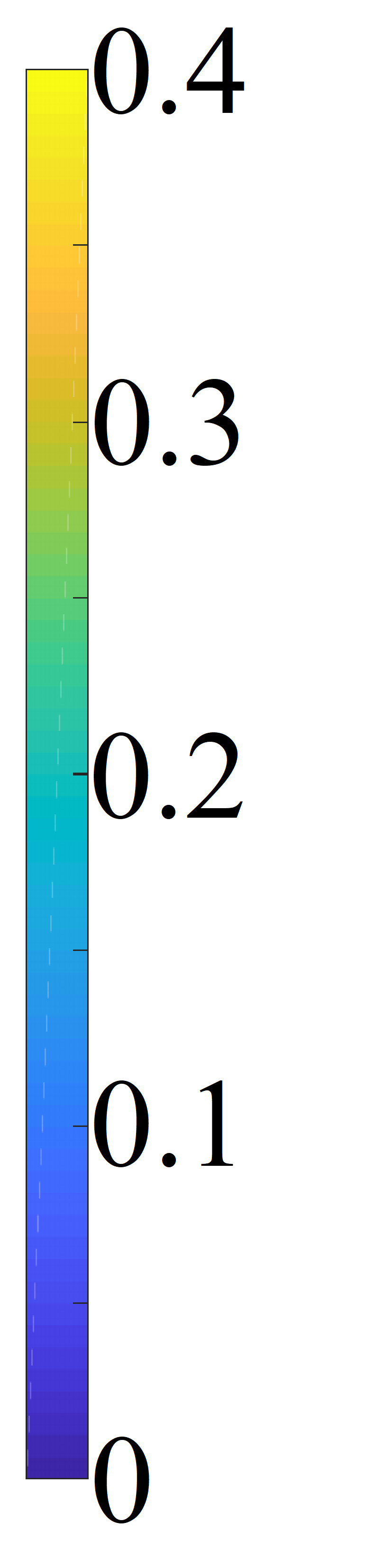} 
\end{tabular}
\end{center}
\caption{Prediction maps of chlorophyll-a content by GP, VHGP and WGP: predictive mean (top row), variance (middle) \blue{and coefficient of variation -ratios- (bottom) are represented.}}
\label{fig:meanchla}
\end{figure}

For illustration purposes, we show in Fig.~\ref{fig:meanchla} the prediction maps, confidence intervals and the ratio between them. The proposed WGP predicts in general lower values of Chl-a content, mostly in the center and eastern region of the scene, and provides a lower uncertainty in the whole image. On the contrary, GP and VHGP obtained low uncertainty only in the western area of the scene and high uncertainty on the central region. Actually, when looking at the ratio between the predictive mean and variance (third row), WGP obtained lower and more homogeneous results. \blue{The lower the ratio, the better the results are. We masked the ratio %out poor predictions 
by fixing a threshold of 20\%: GP provided only 52.21\% of the predictions below this threshold, VHGP yielded a 69.72\%, and WGP fulfilled this error for the whole image.}

\iffalse
\blue{These three methods allow for measuring the uncertainty of the results for a better analysis of the previous predictions. 
Fig. \ref{fig:stdchla} shows the uncertainty regarding Fig. \ref{fig:meanchla}. Exists a big difference between the methods applied. WGP obtained low uncertainty in whole image. As a consequence, the WGP image is blue in its entirety. On the contrary, GP and VHGP obtained low uncertainty only in western area of the image (blues and green tones) and high uncertainty on the center of the image, zone previously mentioned due to the minor chlorophyll-a content predicted by the other method WGP. \textit{The less uncertainty of the WGP method was reflected by the minor value of the standard deviation on Table \ref{tab:res2}.}}

\blue{Fig.~\ref{fig:coefchla} shows the ratio of the standard deviation (Fig.~\ref{fig:stdchla}) to the mean of the predictions obtained (Fig.~\ref{fig:meanchla}).  This ratio known as a relative standard deviation (RSD), allow to express the relationship as a percentage. This approach can compare the results non-dimensionally. Again, WGP obtained blue tones and GP and VHGP obtained heterogeneous results. }

\blue{\textit{Generally, RSD results with percentages above 20\%, should be considered with caution...}. In this context, Fig.~\ref{fig:maskchla} shows in yellow color, the pixels that fulfil the prior premise. GP method achieved just the 52.21\% of the pixels of the image while VHGP achieved the 69.72\% and WGP achieved the entire of the image.  }
\fi

\subsection{Experiment 3: Causal inference in geosciences} \label{sec:experiment3}

Establishing causal relations between random variables from observational data is perhaps the most important challenge in today's Science. In remote sensing and geosciences this is of special relevance to better understand the Earth's system and the complex interactions between the involved processes. % Answering causal questions may have deep societal, economical and environmental implications. %~\cite{Walther02,Adam11}. 

\begin{table}[t!]
\begin{center}
\caption{Problems and causal direction for the geoscience problems in the CEP database.\label{tab:cep}}
\setlength{\tabcolsep}{2pt}
\small
\begin{tabular}{|l|l|l|l|}
\hline
id & $x$ & $y$ & Cause \\
\hline
01  &  Altitude  &  Temperature  &  $\rightarrow$ \\
02  &  Altitude  &  Precipitation  &  $\rightarrow$ \\
03  &  Longitude  &  Temperature  &  $\rightarrow$ \\
04  &  Altitude  &  Sunshine hours  &  $\rightarrow$ \\
20  &  Latitude  &  Temperature  &  $\rightarrow$ \\
21  &  Longitude  &  Precipitation  &  $\rightarrow$ \\
42  &  Day of the year  &  Temperature  &  $\rightarrow$ \\
43  &  Temperature at $t$  &  Temperature at $t+1$  &  $\rightarrow$ \\
44  &  Pressure at $t$  &  Pressure at $t+1$  &  $\rightarrow$ \\
45  &  Sea level pressure at $t$  &  Sea level pressure at $t+1$  &  $\rightarrow$ \\
46  &  Relative humidity at $t$  &  Relative humidity at $t+1$  &  $\rightarrow$ \\
49  &  Ozone concentration  &  Temperature  &  $\leftarrow$ \\
50  &  Ozone concentration  &  Temperature  &  $\leftarrow$ \\
51  &  Ozone concentration  &  Temperature  &  $\leftarrow$ \\
72  &  Sunspots  &  Global mean temperature  &  $\rightarrow$ \\
73  &  CO2 emissions  &  Energy use  &  $\leftarrow$ \\
77  &  Temperature  &  Solar radiation  &  $\leftarrow$ \\
78  &  PPFD  &  Net Ecosystem Productivity  &  $\rightarrow$ \\
79  &  NEP  &  Diffuse PPFDdif  &  $\leftarrow$ \\
80  &  NEP  &  Diffuse PPFDdif  &  $\leftarrow$ \\
81  &  Temperature   &  Local CO2 flux, BE-Bra  &  $\rightarrow$ \\
82  &  Temperature   &  Local CO2 flux, DE-Har  &  $\rightarrow$ \\
83  &  Temperature   &  Local CO2 flux, US-PFa  &  $\rightarrow$ \\
87  &  Temperature  &  Total snow  &  $\rightarrow$ \\
89  &  root decomposition  &  root decomposition (grassland)  &  $\leftarrow$ \\
90  &  root decomposition  &  root decomposition (forest)  &  $\leftarrow$ \\
91  &  clay content in soil  &  soil moisture  &  $\rightarrow$ \\
92  &  organic carbon in soil  &  clay cont. in soil (forest)  &  $\leftarrow$ \\
93  &  precipitation  &  runoff  &  $\rightarrow$ \\
94  &  hour of day  &  temperature  &  $\rightarrow$ \\
\hline
\end{tabular}
\end{center}
\end{table}

In this experiment we used Version 1.0 of the CauseEffectPairs (CEP) collection\footnote{\url{https://webdav.tuebingen.mpg.de/cause-effect/}}. The database contains 100 pairs of random variables along with the right direction of causation (ground truth). Data has been collected from various domains of application, such as biology, climate science, health sciences and economics, to name a few~\cite{Mooij16jmlr}. We conducted experiments in 28 out of the 100 pairs that contain one-dimensional variables and that are related to geosciences and remote sensing: problems involving carbon and energy fluxes, ecological indicators, vegetation indices, temperature, moisture, heat, etc. We summarize the involved variables in Table~\ref{tab:cep}. 

\begin{figure}[h!] 
\vspace{-0.3cm}
\centerline{\includegraphics[scale=.4]{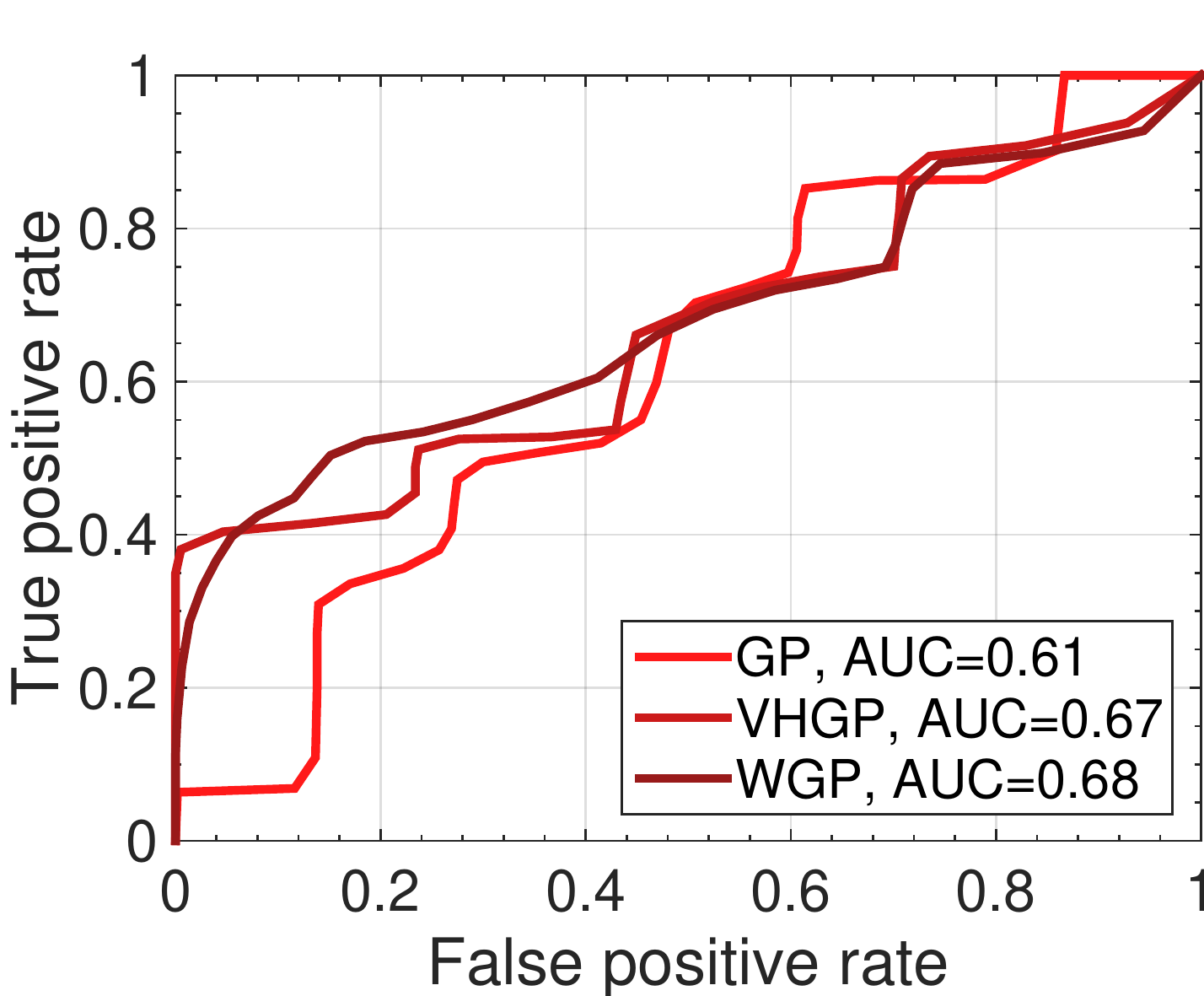}}
%\centerline{\includegraphics[width=.5\columnwidth]{./images/CEP_ROC_averaged100.pdf}\includegraphics[width=.5\columnwidth]{./images/CEP_PR_averaged100.pdf}}
\caption{ROC for the identification of cause-effects in a subset of remote sensing and geosciences pairs using GP (AUC=0.61), VHGP (AUC=0.67) and the proposed WGP (AUC=0.68).\label{res:causal}}
\vspace{-0.4cm}
\end{figure}

We build upon the framework proposed in~\cite{Hoyer08} to derive cause-effect relations from pairs of random variables\footnote{Given two variables, identifying which is the cause and which one is the effect requires adopting (strong) assumptions. For example, it is assumed the absence of `confounding factors' that may drive both variables, that `selection bias' is not present so the observed variables should be representative of the causal relationship, or that feedback loops are not present either~\cite{Pearl2000}.}. The method decides about the causal direction based on the independence of the residuals with respect to the causal mechanism. %(obtained after fitting a regression model in the forward and inverse directions). 
{Notationally, two regression models $\hat{y}=f(x)$ and $\hat{x}=g(y)$ are developed trying to estimate one variable from the other, and then two Hilbert Schmidt Independence Criterion (HSIC)~\cite{Gretton05} terms between the residuals $n_f=\hat{y}-f(x)$ (or $n_b=\hat{x}-g(y)$) and the corresponding potential cause $x$ (or $y$), respectively.} The causal direction score is defined here as the difference in test statistic between both models~\cite{Mooij16jmlr}. For the regression models, we run standard GPs, VHGP and the proposed WGP, and measured independence with HSIC, which has been widely used in remote sensing as well for feature selection and dependence estimation~\cite{campsvalls09grsl}. %,Camps-VallsTLM10}.
Figure~\ref{res:causal} shows the receiver operating curves (ROCs) for all GP models in the case of using a limited number of points $n=1000$ per problem. It can be noticed that all methods perform above chance, and that WGP performs better than the rest in area under the ROC.

%%%%%%%%%%%%%%%%%%%%%%%%%%%%%%%%%%%%
%%%%%%%%%%%%%%%%%%%%%%%%%%%%%%%%%%%%
%%%%%%%%%%%%%%%%%%%%%%%%%%%%%%%%%%%%
\section{Conclusions}
\label{sec:conc}

We introduced warped GPs for remote sensing and geoscience applications involving parameter estimation and causal inference from bivariate problems. WGP regression is able to learn an optimal transform. The introduced model generalizes standard GPs and allows to work with both non-Gaussian processes and non-Gaussian noise. Improved results in terms of accuracy were attained, more credible and tighter confidence intervals, and the advantage of deriving an optimal transform for further use. %We illustrated the advantage of inferring the function for the estimation of bio-physical parameters and causal inference from empirical data.
Results in several experiments suggest that WGPs are a solid approach to both quantitative and qualitative remote sensing data analysis because of the good accuracy and explanatory capabilities. Ongoing work is tied to learning both non-parametric functions following~\cite{Lazaro12warp} on a full Bayesian treatment of warping. %, and optimal invariant transformations from empirical data.

%\blue{NETEJAR EL BIB I ELIMINAR REFERENCIES; EN UNA LETTER NO MES DE 15! }

\bibliographystyle{IEEEbib}
\bibliography{wgp}

\end{document}